# Effective Slot Filling Based on Shallow Distant Supervision Methods

Benjamin Roth   Tassilo Barth   Michael Wiegand
Mittul Singh   Dietrich Klakow
Spoken Language Systems
Saarland University
D-66123 Saarbrücken, Germany
`lsv_trec_qa@lsv.uni-saarland.de`

## Abstract

Spoken Language Systems at Saarland University (LSV) participated this year with 5 runs at the TAC KBP English slot filling track. Effective algorithms for all parts of the pipeline, from document retrieval to relation prediction and response post-processing, are bundled in a modular end-to-end relation extraction system called *RelationFactory*. The main run solely focuses on shallow techniques and achieved significant improvements over LSV's last year's system, while using the same training data and patterns. Improvements mainly have been obtained by a feature representation focusing on surface skip n-grams and improved scoring for extracted distant supervision patterns. Important factors for effective extraction are the training and tuning scheme for distant supervision classifiers, and the query expansion by a translation model based on Wikipedia links. In the TAC KBP 2013 English Slotfilling evaluation, the submitted main run of the LSV *RelationFactory* system achieved the top-ranked $F1$-score of 37.3%.

## 1 Introduction

The English slot filling task of TAC KBP requires participants to extract relational information about query entities of type *person* or *organization* from a large text corpus. At the center of the TAC KBP slot filling task lies the relation detection task, however, steps like document retrieval, finding and disambiguating potential query or answer matches can also have a significant impact on performance. Since TAC KBP slot filling is formulated by stating a well-defined information need, it is designed to shed light into which approaches and steps in a pipeline are most beneficial to solving a query-driven relational extraction task.

The Spoken Language Systems at Saarland University (LSV) 2013 slot-filling system *RelationFactory* is based on the architecture of the LSV 2012 slot filling system [Roth et al., 2012]. *RelationFactory* is based on the driving principles of developing a modular and easily extensible distant supervision relation extractor, making use of shallow textual representations and features. In this paper, we give an overview of the general architecture of the system, and describe novel components and additional evaluations. We will only briefly sketch the components already described in Roth et al. [2012], to which the interested reader is referred.

The paper is structured as follows: Section 2 describes the general system design and discusses each single component in turn. Section 3 discusses the results obtained and provides a detailed per-component evaluation. We give a brief discussion and conclusion in Sections 4 and 5.

## 2 Pipeline and Component Descriptions

The workflow and architecture of *RelationFactory* as illustrated in Figure 1 is based on our last year's system [Roth et al., 2012]. Beyond the changes that we implemented to account for the new requirements in the 2013 task definitions (e.g. justification offsets, `per:title` slots specific to affiliations), we improved the performance of the distant supervision SVM classifier, included a distant supervision-based pattern module in the main run, and included experimental modules (dependency patterns, Wikipedia-based validator) in additional runs. Our pipeline is a two-stage pipeline with a candidate generation stage (consisting of document retrieval and sentence filtering based on named-entity type checking and query matching) and a candidate validation stage (see Figure 1). The candidate validation stage consists of several modules that decide (typically based on the relational context) whether a candidate indeed expresses the relation or not.

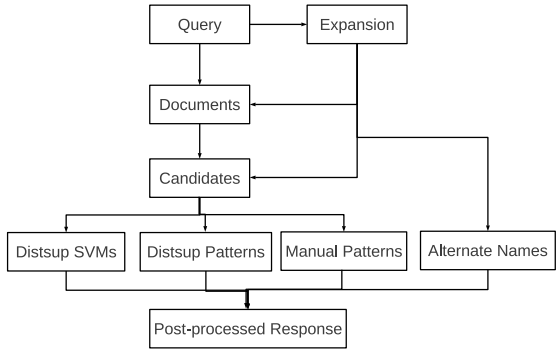

Figure 1: Data-flow of the *RelationFactory* relation extraction system. The modules before post-processing the response form the *candidate validation* stage.

We used the same training data and hand-crafted surface patterns as in Roth et al. [2012], all improvements with respect to last years' system are due to advances on the modeling and algorithmic side. The main improvements, which are included in the main run, stem from an improved and consolidated feature representation and a recently developed distant supervision pattern scoring scheme.

## 2.1 Query Expansion, Retrieval and Candidate Matching

The name of a TAC KBP query entity that is provided is expanded by a translation model based on Wikipedia anchor text, inspired by work on cross-language information retrieval [Roth and Klakow, 2010]. The advantage of using anchor text rather than e.g. Wikipedia redirects is that anchor text captures a wide range of variations, as they occur in actual sentences. In order to avoid translations to surface forms that mainly denote other entities, we only retain query expansions for which the most frequently co-occurring Wikipedia page is the same as for the original query name (*link-back requirement*). Example output for this type of query expansion is displayed in Figure 2.

For queries of type *organization*, additional expansions are generated by augmenting the original name by common suffixes of business forms (taken from a list in Wikipedia; e.g. *Ltd*, *Corp*). For queries of type *person*, additional expansions are obtained by adding the last name of a person only (this expansion is, however, not used for document retrieval). We use the expansions for retrieval and matching directly and do not use any other entity linking or disambiguation strategies as entity linking is not the focus of this work.

We retrieve documents by using the original query name and a query expansion, selected by high pointwise mutual information w.r.t. the query. Sentences are tagged using a state-of-the sequence perceptron named-entity tagger [Chrupała and Klakow, 2010]; non-standard named-entity types relevant for certain relations are matched using lists of respective Freebase types. Sentences where a query or one of its expansions matches, and that have a named-entity tag of a potential slot filler type, are passed to the prediction components for validation. More details on query name expansion, retrieval and candidate matching can be found in last years' system description [Roth et al., 2012].

## 2.2 Distant Supervision SVM Classifiers

The most important candidate validation component (see Figure 1), both in terms of stand-alone F1-score, as well as F1 contribution in the ablation analysis, is the set of distantly supervised per-relation SVM classifiers.

| Original query | Wikipedia link anchor text expansions | Per: last name / Org: suffixes |
|---|---|---|
| Ali Akbar Khan | Utd. Ali Akbar Khan, Ustad Ali Akbar Khan | Khan |
| Adam Gadahn | Azzam the American, Adam Yahiye Gadahn | Gadahn |
| Nancy Kissel | Murder of Robert Kissel, Robert Kissel | Kissel |
| DCNS | Direction des Constructions Navales, DCN, ... | DCNS Ltd, DCNS Corp, ... |
| STX Finland | Kvaerner Masa Yards, Aker Finnyards, ... | STX Finland Ltd, ... |

Figure 2: Examples of query expansions. The expansion of *Nancy Kissel* is an example of a wrong expansion to thematically related entities. The vast majority of query expansions is, however, beneficial.

### 2.2.1 Training Data

The training data is the same as used in Roth et al. [2012]. Distant supervision argument pairs are obtained by mapping Freebase relations to TAC relations and by matching hand-crafted seed patterns against the TAC 2009 text collections. This way we obtain two sets of seed pairs. We use a maximum of $10k$ argument pairs per relation for each of the two sets of seed pairs. These pairs are then matched against the TAC 2009 text corpora, and a maximum of 500 sentences per pair are used as training data.

### 2.2.2 Feature Set

In comparison to last years' system we use a rather minimalistic feature set. We do not include most of the previously used features (e.g. argument features, distance features, Brown cluster features, etc.) but only model context with token n-gram-based features. When using token n-grams, we found it essential to mark whether the query (referred to as $ARG1$) or the slot filler ($ARG2$) comes first. Additionally, including *sparse* (or *skip*) n-grams, where tokens in the middle of the n-gram were wildcarded, increased performance. For the context between $ARG1$ and $ARG2$, we use n-grams up to length 3 and skip-n-grams of length 3 and 4. We model the left and right context outside the arguments with n-grams up to length 3 (including the corresponding wildcarded argument). Figure 3 shows examples of extracted features for a candidate instance.

### 2.2.3 Aggregate Training and Parameter Tuning

We train one binary support vector machine for each of the relations using the distant supervision matches for that relation as positive data and the matching contexts for all other relations as negative data.[1] We use $SVM^{light}$[2] as the classification toolkit. Two effective mechanisms to increase distant supervision training performance are employed: We refer to the first as *aggregate training*, the second is *global parameter tuning*.

**Aggregate training.** Distant supervision training data contains noisy false positive sentence matches of pairs in the knowledge base. Min et al. [2012] approach this problem by a method called label-refinement: the matching sentences are taken as instances and a classifier model is trained on them. In a second iteration, the training data is classified (relabeled) by this model, and the final model is trained using the resulting *refined* labels. This relabeling should reduce ambiguity in the training data, and enforce common patterns for a relation found in the first iteration.

In our experiments, we found a another extremely simple method, which we call *aggregate training*, to be more effective. Here, instead of treating each distant supervision match (i.e. each individual sentence in which an entity pair matches) as a single training example, we

---
[1] If the same feature vector happens to occur more than once in the training data, and is labeled both as positive and negative, those instances of the feature vector which are labeled as negative are removed from the training data.
[2] http://svmlight.joachims.org/, Joachims [1999]

**Relation:** `per:origin(Adam Gadahn, U.S.)`

**Candidate sentence:** *One Pakistani intelligence official said he is Adam Gadahn, a California native and the first U.S. citizen to be charged with treason in 52 years.*

**Feature examples:**
BETWEEN_NGRAM#ARG1>#,>
OUTSIDE_NGRAM#ARG2>#citizen>#to>
SKIP_NGRAM#native>###first>

Figure 3: Examples of extracted features. Each feature is first marked with the feature group it belongs to (n-gram between or outside the arguments, skip-n-gram), followed by the token sequence of the n-gram, using # as a separator. Each token is marked to indicate whether the slot filler comes left (<) or right (>) of the query.

group all sentences per entity pair, extract the features, sum the feature counts of all these sentences and normalize the feature vector for that pair so that the highest feature has weight 1.0.

This scheme greatly increases training speed. Moreover, it mitigates a potential skew in the training data by features that are highly correlated with a distant supervision pair (e.g. the word *president* for the pair *Barack Obama, Michele Obama*), but not with the respective relation (`per:spouse`), which might lead the classifier to model frequently matching argument pairs rather than generalizing to the relation. More investigations into the effectiveness of this training scheme are left for future work.

**Global parameter tuning.** Tuning the cost-factor by which training errors on positive examples outweigh errors on negative examples (also called $j$-parameter in SVM$^{light}$) has often been observed as crucial to performance. Moreover, experimental results suggest that simple misclassification cost tuning is superior to multi-instance learning in many settings [Ray and Craven, 2005] including relation extraction [Bunescu and Mooney, 2007].

We therefore trained three SVM configurations for each relation by setting the $j$-parameter to 0.1, 1.0 and 10.0, respectively. We found that the best local parameter choice (i.e. the parameter settings that produce best per-relation F1-scores) does not necessarily correspond to an optimal global (micro-average) F1-score: For example, for relations with a low precision over the whole recall range (e.g. due to errors in a previous tagging step), increasing the individual F1-score by increasing recall may have a negative overall effect. Likewise, for relations with an above average precision, it may be beneficial for overall performance to score more instances as positive than tuning for individual F1-score may result in.

To avoid these problems that arise by individually maximizing per-relation F1-scores, we use a greedy procedure to tune the per-relation $j$-parameters in order to optimize global F1-score, instead. Algorithm 1 shows the pseudo-code of the global parameter-tuning. We use $\mathcal{R}$ to denote the set of relations, $j(r)$ a choice of parameter for a particular relation $r \in \mathcal{R}$, *evaluate()* a function returning the global F1-score for the current choices of $j(\cdot)$, and *evaluate($j(r)\backslash j$)* the global F1-score with a particular $j(r)$ replaced by $j$. The parameters are tuned with respect to performance on earlier TAC KBP slot filling queries (years 2009–2012).

### 2.2.4 Prediction

While training is done on an aggregate level, prediction is done on each candidate sentence independently. The per-sentence prediction is motivated as in TAC KBP, the task is not to find pairs that *likely* belong into the knowledge base (e.g. by indirect correlations), but to find pairs that *justifiably* belong into the knowledge base (i.e. actual sentences must express the relations). An answer is returned if at least one

**Algorithm 1** Global parameter tuning. The second loop over the relations can be executed iteratively (in our setting it was executed twice).

>  **for** $r \in \mathcal{R}$ **do**
>     $j(r) \leftarrow 0.1$
>  $f_1 \leftarrow evaluate()$
>  **for** $r \in \mathcal{R}$ **do**
>     **for** $j \in \{0.1, 1.0, 10.0\}$ **do**
>        $\hat{f}_1 \leftarrow evaluate(j(r)\backslash j)$
>        **if** $\hat{f}_1 > f_1$ **then**
>           $f_1 \leftarrow \hat{f}_1$
>           $j(r) \leftarrow j$

candidate sentence with it is classified as *true*.[3]

## 2.3 Distant Supervision Patterns

As a second distant supervision component besides the SVM classifiers, we include simple intertext patterns (the lexical token sequence between the arguments) that are scored according to frequency in the distant supervision data, and two combined noise reduction methods to suppress the influence of false positive matches (see Roth et al. [2013] for an overview of approaches to distant supervision noise reduction).

The pattern scoring follows the method of Roth and Klakow [2013] and combines for each pattern the count of the respective relational topic on the training data $n(pat, topic(r))$ and the score of a discriminatively trained Perceptron model $P(r|s, \theta)$. The overall scoring function further includes the relative frequency of a pattern for a specific relation $\frac{n(pat,r)}{n(pat)}$ and the perceptron probability $P(NIL|s, \theta)$ of the pattern to express no relation. It is denoted by:

$$\frac{0.5 \cdot n(pat, topic(r))}{n(pat)} + \frac{0.5 \cdot n(pat, r) \cdot P(r|s,\theta)}{n(pat) \cdot (P(r|s,\theta) + P(NIL|s,\theta))}$$

The scoring function provides scores in the interval between 0.0 and 1.0. We use the same global parameter tuning method as for the distant supervision SVM classifiers to find score thresholds on the intertext patterns (see Algo-rithm 1). We tune thresholds on the score levels $0.1, 0.3, 0.5, 0.7$ and $0.9$.

---
[3]For single slot relations, only the answer with the highest classifier regression score is returned.

## 2.4 Influence of Hand-Crafted Patterns

In TAC KBP, the task is defined by a human readable task description, mostly independent of restrictions on the kind of methods to be used. The mapping of that task description into an automatic system always requires human effort. The most popular approaches to capture that human translation step is by providing hand-crafted patterns or manually establishing mappings to knowledge-bases such as Wikipedia infoboxes or Freebase.

In order to keep the effort to pattern writing minimal, in our system we restricted the patterns to plain sequences of tokens with a general placeholder (* denoting 1 to 4 tokens) and did not use syntactic patterns that would require linguistic expertise. We use the same patterns as in our 2012 system (the patterns for `per:member_of` and `per:member_of` were merged for the 2013 system). Although we found it generally to be less effort to write down a few token sequences than to identify the corresponding relational correspondence in Freebase (especially since certain sequences follow directly from the examples and definitions of the task description), it is interesting to quantify the influence of the hand-crafted patterns in our system.

We therefore compare the performance of our hand-written patterns to the reported scores of hand-written patterns in the NYU 2012 system [Min et al., 2012]. In the NYU 2012 system there are three modules with dedicated hand-crafted patterns: A so-called *local patterns* module, that includes short patterns similar to ours, and two *bootstrapped patterns* modules that take additional dedicated hand-crafted patterns as an input and iteratively add new patterns, based on corpus co-occurrences. The NYU pattern bootstrapping modules use hand-crafted patterns both based on token sequence and syntactic paths.

Table 1 shows the performance of the NYU hand-written pattern modules and our hand-written pattern module for the TAC 2012 task.

| System / Pattern Component | Precision | | Recall | | F1-Score | |
|---|---|---|---|---|---|---|
| NYU / local patterns | 47.4 | | 9.3 | | 15.6 | |
| NYU / bootstrapped linear | 59.2 | | 4.6 | | 8.5 | |
| NYU / bootstrapped dependency | 54.8 | | 3.7 | | 6.9 | |
| LSV / token sequence | 43.1 | (49.0) | 8.0 | (8.4) | 13.5 | (14.3) |

Table 1: Comparison of the NYU hand-crafted pattern modules and the hand-crafted pattern component used in our system (LSV), on the 2012 task. For the LSV system we give the exact evaluation of the 2012 system, and in brackets the *anydoc* and *lowercase* evaluation of the currently used system.[4]

It should be noted that performance of a particular module is also affected by other factors such as retrieval, argument matching and postprocessing. The performance of the hand-written patterns in our system roughly corresponds to that of the NYU hand-written local patterns component.

### 2.5 Alternate Names

Entity pairs of the relation `alternate names` can be predicted by any of the validation components such as the SVM classifier or a pattern matcher. Additionally, we include a dedicated component that explicitly returns a slot filler for `per:alternate_names` or `org:alternate_names` if an expression returned by our query expansion (see Section 2.1) matches in one of the retrieved documents.

### 2.6 Postprocessing and Redundancy Removal

Postprocessing and redundancy removal are based on mapping the answers to normal forms based on Wikipedia link translations and lowercasing as in the 2012 system [Roth et al., 2012]. Additionally, due to the changed task description for the `per:title` relation, we included job titles multiple times if they co-occurred with different organization names, and the co-occurrence was licensed by a pattern.[5] However, the special treatment of `per:title` decreased overall performance. We do not use any cut-off on the number of returned answers per slot.

### 2.7 Modules not Included in the Main Run

**PRIS Syntactic Patterns.** We implemented a module to match the dependency patterns provided by the PRIS team [Li et al., 2012]. Thus we wanted to test whether dependency patterns may help to improve performance in our pipeline. Due to the many degrees of freedom to incorporate those patterns into a relation extraction system, we cannot guarantee that our module makes the best (or even correct) use of the provided patterns.

**Wikipedia-Based Validator.** This module runs the relation extraction pipeline on an additional Wikipedia text dump and uses the slot fillers obtained this way to validate candidates retrieved from the TAC corpora.

## 3 Results and Evaluation

### 3.1 Submitted Runs

Table 2 gives an overview over the submitted runs. They are characterized as follows:

- **lsv1 (Main Run):** In this run, only *fast* validation components are used, this means especially no syntactic analysis and no query-specific analysis of an additional Wikipedia dump. The fast components are the SVM classifier, the distant supervision patterns, the hand-written patterns, and the alternate names expansion module.

---

[4] *anydoc* is an option in the TAC scorer for scoring independently of the reported document id, *lowercase* is an option for case insensitive scoring. When running the current systems the relations `per:employee_of` and `per:member_of` are merged both for prediction and evaluation.

[5] The list of patterns was compiled from high-frequency context patterns between entities of type [PERSON] and [ORGANIZATION].

| run id | run type | P | R | F1 |
|---|---|---|---|---|
| lsv1 | fast | 42.5 | 33.2 | 37.3 |
| lsv2 | precision | 50.9 | 25.9 | 34.3 |
| lsv3 | all | 36.9 | 36.6 | 36.8 |
| lsv4 | recall | 35.1 | 37.8 | 36.4 |
| lsv5 | all shallow | 38.1 | 35.8 | 36.9 |

Table 2: Official scores on 2013 runs submitted by team LSV.

- **lsv2:** Only modules are included that produced *high precision* on the 2012 development data. This includes most components of *lsv1*, but not the SVM classifier. Additionally the syntactic patterns are included in this run.

- **lsv3:** This contains *all validation components* with standard configuration. It includes all components from *lsv1* and *lsv2*, and the Wikipedia-based validator.

- **lsv4:** This is a *high-recall* run. In addition to the components of *lsv3*, the entity expansion is relaxed (ambiguous expansions are allowed), and `per:employee_or_member_of` slots are inferred from predicted `per:title` slots (if a title is predicted, then a co-occurring organization name may be returned).

- **lsv5:** This is a run that exclusively comprises *shallow* components (i.e. no syntactic analysis). It corresponds to *lsv1* together with the Wikipedia-based validator.

Interestingly, the *fast* run (*lsv1*), that only extracts surface-level features and matches linear patterns, is the best performing in terms of $F1$ score. Increasing the precision by concentrating on high-precision modules as well as increasing recall by merging responses from more modules did not have an overall positive effect. It remains for future work to analyze whether additional improvements can be achieved by a more principled module combination scheme (rather than simply merging the responses).

### 3.2 Single Component Analysis and Ablation Analysis

Table 3 shows the performance of the single components and the merge of their responses. In order to show how complementary single components are with respect to the other components, Table 4 gives an ablation analysis on the best-performing run (*lsv1*). Some observations on the performance of single components:

- **Alternate Names.** The inferred `alternate_names` slot fillers from the query expansion are high-precision. Although concerned with only two relations, this component gives a F1 gain of 1.9% on top of the other components.

- The **hand-crafted patterns** provide high-precision responses, but have relatively low recall for a component modeling all relations. They are considerably complementary w.r.t. the other components (+4.1% F1).

- **Distsup patterns.** The patterns induced from the distant supervision data provide good-precision responses with good recall. They capture information not modeled by either the SVM classifiers or the hand-crafted patterns (+4.1% complementary F1 gain).

- **PRIS syntactic patterns.** The dependency patterns have good precision, but are slightly behind plain surface patterns in our experiments.

- **Distsup SVM classifier.** The SVM classifiers are the strongest relation validation component in our system, both in terms of single performance as well as complementary F1 gain (+6.4% F1).

- **Wikipedia-based validator.** This is the component with the lowest precision, since apart from candidate generation (query

| Component | P_single | P_merge | R_single | R_merge | F1_single | F1_merge |
|---|---|---|---|---|---|---|
| Alternate names | 54.2 | – | 1.8 | – | 3.4 | – |
| Hand-crafted patterns | 50.2 | 50.4 | 10.3 | 12.0 | 17.1 | 19.4 |
| Distsup Patterns | 42.7 | 53.5 | 15.6 | 21.9 | 22.9 | 31.0 |
| PRIS syntactic patterns | 39.0 | 50.4 | 9.6 | 25.6 | 15.4 | 34.0 |
| Distsup SVM classifier | 34.7 | 40.5 | 23.6 | 34.3 | 28.1 | 37.2 |
| Wiki validator | 20.8 | 36.9 | 8.1 | 36.7 | 11.7 | 36.8 |
| +inferred `per:title` affiliations | – | 36.0 | – | 37.7 | – | 36.8 |
| +relaxed query expansion | – | 35.1 | – | 37.8 | – | 36.4 |

Table 3: Performance of single component and merged component responses. Components are sorted by precision. The last two components cannot be evaluated in isolation: the component "inferred `per:title` affiliations" operates on an already existing response, the component "relaxed query expansion" influences the number of candidates fed into validation components. These two components are evaluated in conjunction with the merge of all components up to the Wikipedia-based validator.

| Component | P | R | F1 | F1 gain |
|---|---|---|---|---|
| main run | 42.5 | 33.2 | 37.3 | |
| −Query expansion | 41.1 | 17.5 | 24.5 | +12.8 |
| −Distsup SVM classifier | 53.3 | 21.8 | 30.9 | +6.4 |
| −Distsup patterns | 39.6 | 28.6 | 33.2 | +4.1 |
| −Hand-crafted patterns | 38.2 | 29.5 | 33.2 | +4.1 |
| −Alternate names | 41.1 | 31.0 | 35.4 | +1.9 |
| −Redundancy removal | 41.4 | 33.2 | 36.8 | +0.5 |
| −Multiple `per:titles` | 44.0 | 33.0 | 37.7 | −0.4 |

Table 4: Precision, Recall and F1-score of the main run configuration when removing single components (one at a time), as well as the F1 gain contributed by the respective component on top of the other components. Components are sorted by complementary F1 gain.

matching, tagging) only overlap with answers from Wikipedia is checked. It is interesting to note that while this component obtained high precision in our internal development benchmarks, precision was rather low for the official submitted run.[6]

- **Inferred `per:title` affiliations.** Inferring `per:employee_or_member_of` from predicted `per:title` relations had a minimal effect on the precision/recall ratio.

- **Query expansion** and **Relaxed query expansion.** It is important to note that query expansion has a high effect on overall performance, contributing a F1 gain of 12.8%. This is due to the greatly positive effect on recall, while almost not harming precision. Query expansion plays a role in both document retrieval and query matching. It seems necessary not to overgenerate, as predicting more ambiguous aliases (no link-back requirement, see Section 2.1) increased recall but had negative effect on F1-score.

- **Redundancy removal.** Removing redundant slot fillers using Wikipedia anchor text had a slightly beneficial effect on overall F1.

- **Multiple `per:title`s.** On the other hand, trying to cluster predicted `per:title`s by their affiliations (see Section 2.6) was detrimental to performance.[7]

## 4 Discussion: Shallow vs. Deep Analysis

In our main run no deep linguistic analysis, such as dependency parsing, was used. Merely named-entity tagging was used to identify slot filler candidates – all the features and patterns operate directly on the surface level. When developing the *RelationFactory* KBP system, we kept experimenting with more linguistically motivated representations but found that they did not provide any (substantial) gain over representations derived directly from the surface forms. While syntactic representations (especially dependency relations) provide a metaphor every researcher in the field is accustomed to when speaking about relational representations, our observations suggest that taking one step back from the dependency view may clear the sight to more central aspects of certain information extraction tasks.

Apart from purely practical advantages of a shallow approach (e.g. faster code that is easier to maintain, applicability in low-resource settings), there are also more considerations:

- **Contextual cues.** Words or word sequences that do not express the relation but provide topical information and may disambiguate a relational expression are naturally included in a shallow feature representation. A dependency analysis, however, aims to strip off those cues.

- **Micro-structures without content words.** Chan and Roth [2011] observe that in ACE 80% of the mention pairs in a relation do fall in a pattern type where the relation is not explicitly expressed by a content word. The four pattern types they identify are *Premodifier* (e.g. [the [Seattle] Zoo]), *Possessive* (e.g. [[California's] Governor]), *Preposition* (e.g. [officials] in [California]) and *Formulaic* (e.g. [Medford], [Massachusetts]).

- **Parsing errors.** While syntactic parses may be accurate for short distance dependencies, which also can be easily captured by surface patterns, for longer distances the dependency accuracy significantly decreases [McDonald and Nivre, 2007].

---

[6]In development, the Wikipedia-based validator achieved an *'anydoc'* precision of 36% on 2012 data. It is to be expected that evaluation independently of the document id results in higher scores – however, the Wikipedia validator was the only module with such a big mismatch between development and submission scores.

[7]The components related to post-processing, *redundancy removal* and *multiple `per:titles`* are part of every run in Table 3 and therefore only separately evaluated in the ablation study (Table 4).

## 5 Conclusion

The LSV 2013 English slot filling system *RelationFactory* is a distant supervision system for query-based relation extraction. It is based on query expansion by anchor text translations, hand-crafted seed patterns and two distant supervision components: one modeling relation prediction as a classification task using Support Vector Machines and shallow features; the other scoring surface patterns by a combination of generative and discriminative distant supervision noise reduction models. A detailed analysis showed that each of the aforementioned components contributed to the overall good performance of the system. Other components that were not included in the main run, such as handwritten dependency patterns or validation by answers found in a Wikipedia text corpus, could not improve on the results achieved with these basic components.

## Acknowledgements

Benjamin Roth is a recipient of the Google Europe Fellowship in Natural Language Processing, and this research is supported in part by this Google Fellowship. Tassilo Barth was supported in part by IARPA contract number W911NF-12-C-0015 and Michael Wiegand by the German Federal Ministry of Education and Research (BMBF) under grant no. "01IC10S01".